\documentclass[10pt,twocolumn,letterpaper]{article}

\usepackage[pagenumbers]{cvpr} 

\usepackage[dvipsnames]{xcolor}

\usepackage{multirow}
\newcommand{\tabincell}[2]{\begin{tabular}{@{}#1@{}}#2\end{tabular}}
\definecolor{cvprblue}{rgb}{0.21,0.49,0.74}
\usepackage[pagebackref,breaklinks,colorlinks,citecolor=cvprblue]{hyperref}

\def\paperID{1991} 
\def\confName{CVPR}
\def\confYear{2024}

\title{Is Conventional SNN Really Efficient? \\ A Perspective from Network Quantization}

\author{Guobin Shen$^{1, 3}$ \ \ Dongcheng Zhao$^{1}$ \ \ Tenglong Li$^{1, 4}$ \ \ Jindong Li$^{1, 4}$ \ \ Yi Zeng$^{1, 2, 3, 4}$\thanks{Corresponding author: yi.zeng@ia.ac.cn}\\
$^1$Brain-inspired Cognitive Intelligence Lab, Institute of Automation, Chinese Academy of Sciences\\ 
$^2$ Center for Excellence in Brain Science and Intelligence Technology, Chinese Academy of Sciences \\
$^3$ School of Future Technology, University of Chinese Academy of Sciences  \\ 
$^4$ School of Artificial Intelligence, University of Chinese Academy of Sciences  \\ 
\texttt{\{shenguobin2021, zhaodongcheng2016, } \\ 
\texttt{litenglong2023, lijindong2022, yi.zeng\}@ia.ac.cn}
}

\begin{document}
\maketitle
\begin{abstract}



Spiking Neural Networks (SNNs) have been widely praised for their high energy efficiency and immense potential. However, comprehensive research that critically contrasts and correlates SNNs with quantized Artificial Neural Networks (ANNs) remains scant, often leading to skewed comparisons lacking fairness towards ANNs. This paper introduces a unified perspective, illustrating that the time steps in SNNs and quantized bit-widths of activation values present analogous representations. Building on this, we present a more pragmatic and rational approach to estimating the energy consumption of SNNs. Diverging from the conventional Synaptic Operations (SynOps), we champion the "Bit Budget" concept. This notion permits an intricate discourse on strategically allocating computational and storage resources between weights, activation values, and temporal steps under stringent hardware constraints. Guided by the Bit Budget paradigm, we discern that pivoting efforts towards spike patterns and weight quantization, rather than temporal attributes, elicits profound implications for model performance. Utilizing the Bit Budget for holistic design consideration of SNNs elevates model performance across diverse data types, encompassing static imagery and neuromorphic datasets. Our revelations bridge the theoretical chasm between SNNs and quantized ANNs and illuminate a pragmatic trajectory for future endeavors in energy-efficient neural computations.

\end{abstract}  
\section{Introduction}
\label{sec:intro}



Spiking Neural Networks (SNNs)~\cite{maass1997networks}, rooted in a computational framework that draws inspiration from biological neural processes, present a compelling counterpoint to the established paradigms of traditional Artificial Neural Networks (ANNs). This modality signifies a substantial divergence from the analog and continuous activation regimes that characterize ANNs. Predicated on an event-driven operational principle, SNNs are inherently poised for reduced energy consumption, a trait that is instrumental in the ongoing evolution of neuromorphic computing platforms~\cite{li2023firefly}. Building on the energy-efficient principles inherent to SNNs, pioneering developments in neuromorphic chips like TrueNorth~\cite{merolla2014million} and Loihi~\cite{davies2018loihi}  have underscored their significant potential for reducing energy consumption in comparison with traditional computing approaches. These platforms capitalize on the efficiency of SNNs to offer optimized solutions for energy-constrained applications, advancing the frontiers of computing technology.

Despite the inherently low power consumption of SNNs, their deployment in edge computing devices such as mobile phones and wearable technology remains notably infrequent. Despite progress in neuromorphic technology, modern SNN configurations remain deeply rooted in the methodologies of deep ANNs. These SNN designs typically rely on optimistic energy usage projections that overlook the tangible overheads associated with deploying hardware in real-world environments~\cite{li2023firefly}. This approach can distort the comparison with traditional ANNs, especially when such ANNs have been refined for energy-efficient operation. To address the severe energy and computational constraints, the traditional machine learning domain has introduced a suite of model optimization techniques, including quantization~\cite{liu2021post, lin2021fq, li2022q} and network pruning~\cite{gordon2020compressing, sanh2020movement, han2023enhancing}. These methods effectively reduce the computational and memory requirements. Such adaptability ensures that sophisticated neural models can be deployed even in resource-limited settings with minimal compromise on performance. Conventionally, SNNs are often compared to ANNs that have not undergone similar optimizations~\cite{yao2023spike, zhou2022spikformer, shen2022backpropagation, rathi2020diet}. Although such comparisons may highlight the energy efficiency of SNNs, they do not fairly consider the significant advancements in ANN energy-efficient optimization. Consequently, this prevalent comparative framework fails to substantiate the computational efficacy of SNNs in a comprehensive manner and does not adequately acknowledge the considerable enhancements achieved through contemporary ANN inference optimization methodologies.

In particular, the advent of surrogate gradient methods~\cite{bohte2011error} has precipitated new advancements in the field of deep SNNs, pivoting the focus towards enhancing network performance concurrently with the reduction of network latency~\cite{zhao2023improving,shen2022backpropagation,zeng2023braincog}. Incorporating advanced methodologies has precipitated a notable reduction in latency within deep SNNs, culminating in scenarios where the processing is streamlined to necessitate a mere one or two time steps. This observation raises a compelling inquiry: Does an SNN, functioning within the confines of a single time step, essentially mirror the characteristics of an ANN wherein the activation is quantized? Our analysis scrutinizes the efficiency of SNNs through the prism of neural network quantization, allowing us to elucidate the nuanced discrepancies between SNNs and their quantized ANN counterparts. We undertake a thorough and methodical examination of the computational efficacy of SNNs, with an emphasis on their practical application within real-world contexts, transcending beyond the confines of theoretical computational advantages. Our contribution can be summarized as follows:

\begin{figure*}[htbp!]
  \centering
  \includegraphics[width=0.9\textwidth]{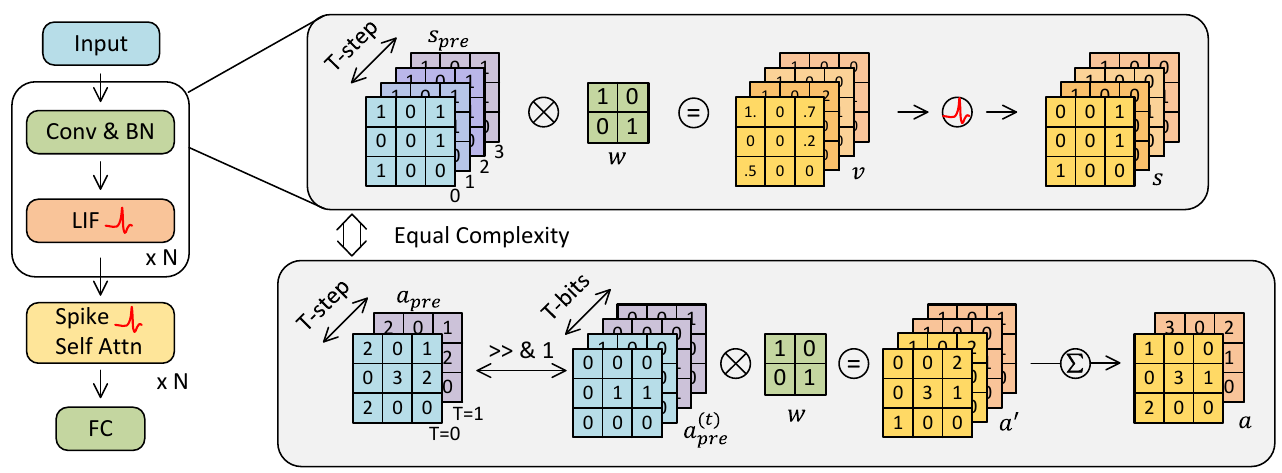}
  \caption{With the same number of feature bits, SNNs and quantized ANNs have the same complexity of representation.} 
  \label{fig:abstract} 
  \vspace{-6mm}
\end{figure*} 

\begin{itemize}
  \item We introduce a unified framework for quantitatively analyzing SNNs and ANNs, demonstrating that an SNN with simulation step \( T \) is comparable to a \( T \)-bit quantized ANN. This emphasizes the association between SNNs and ANNs in terms of computational complexity. 
  \item We define the "Bit Budget" to measure synaptic operation complexity and present two metrics, S-ACE and NS-ACE, to assess computational demand across hardware platforms. Time steps, spike patterns, and weights are taken into account for the computational overhead. These metrics inform the development of the strategy and emphasize the significant impact of model weight quantification on computation and storage
  \item Our validation across varied neural tasks and architectures highlights the potential for our findings to bridge SNN research with broader deep learning advances and guide future explorations in the field.
\end{itemize}

\section{Related Work}
\label{sec:related_work}

Spiking neural networks, heralded as the third generation of neural networks, are engineered to emulate the intricate information-processing mechanism of the human brain. Maass~\cite{maass1997networks} was seminal in delineating the contrast between SNNs and their ANN counterparts, explicitly highlighting the former's proficiency in discrete-time, event-driven processing. This paradigm shift was further catalyzed by the work~\cite{shen2022backpropagation, zhang2021rectified,zhang2018highly,wu2021progressive} and innovative Spiking Transformer architectures~\cite{yao2023spike, zhou2022spikformer}. These models are increasingly becoming cornerstones in domains like image recognition, natural language processing, and robotics control~\cite{zheng2021going, zhu2023spikegpt, shen2023brain}, underscoring the SNNs' inherent ability to execute computationally demanding tasks in an energy-efficient manner, presenting a sustainable alternative to conventional ANNs.

Optimization of neural network efficiency, particularly in terms of computational overhead and resource utilization, is a critical aspect of advancing deep learning technology. Notable techniques such as network pruning and quantization have dramatically diminished the computational burden and storage requirements traditionally associated with neural networks. Pruning strategies streamline network architectures by systematically eliminating superfluous connections and neurons. Han et al.~\cite{han2015deep} illustrated the over-parameterization in deep learning models and optimized them through pruning. This approach evolved, giving rise to structured~\cite{liu2018rethinking, li2016pruning} and unstructured pruning methods~\cite{han2015learning}. Quantization, on the other hand, concentrates on minimizing the precision of the numerical representation of neural weights and activations, substantially saving on both storage and computational intensity. Zhou et al.~\cite{zhou2016dorefa} showcased the potential of quantized neural networks. Extreme cases, like Binary Neural Networks (BNNs), which only use single-bit representations, have also been explored~\cite{courbariaux2016binarized}. Thereon, the domain has witnessed extensive research, focusing on balancing quantization and performance~\cite{nagel2020up}.

Given the intertwined trajectories of SNNs, neuromorphic hardware, and ANNs, there is an evident need for a comprehensive evaluation metric that accounts for practical hardware constraints. However, current measures of the computational complexity of SNNs using the number of synaptic operations (SOPs)~\cite{hu2021spiking} only take into account the effects of spike rate and time step, making it challenging to meet this burden. Our research suggests that SNNs and ANNs can be equated to circuit complexity. Recognizing this equivalence provides an avenue to analyze SNN energy consumption in line with ANNs, necessitating an inclusive energy-efficiency evaluation approach. Embracing techniques from ANNs, especially quantization can further amplify the energy-saving potential of SNNs. Informed by this holistic perspective, we introduce a universal framework. Our approach delves deep into the facets of weight precision, activation function choices, and the unique temporal resolution aspect of SNNs. This framework is designed to bridge theoretical postulations with practical implementations, championing both energy efficiency and computational excellence.

\section{Method}

\subsection{T-step SNNs and T-bit Quantized ANNs}
\label{sec:qnn_vs_snn}

Both SNNs and Quantized ANNs (QANNs) enhance the efficiency of network inference through discrete representations. However, literature rarely discusses the differences and connections between them. We analyze both from a unified perspective.

An SNN layer operates across three temporal phases, delineated in Eq. \ref{lif}. The input spikes ($s_{\text{pre}}$) from the antecedent layer are weighted and transformed into currents that modulate the neuron's membrane potential ($v_{t}$). If this potential exceeds a certain threshold ($v_{th}$), it resets after a spike, encapsulated by the Heaviside step function ($H(\cdot)$), reflecting the all-or-none nature of neuronal firing. $v_{rst}$ is the reset potential, $\tau$ is the time constant, and $w_{j}$ is the synaptic weight for the $j^{th}$ presynaptic neuron.

\begin{equation}
    \begin{aligned}
        v_{t} & = \frac{1}{\tau} v_{t-1} + \sum_{j} w_{j}s_{\text{pre}}, \\
        & \quad \quad \text{if} \ v > v_{th} \ \text{then} \ v \leftarrow v_{rst}, \\
        s &= H(v - v_{th}) 
    \end{aligned}
    \label{lif}
\end{equation}

QANNs mirror SNNs in structure but not in temporality, operationalized through Eq. \ref{quant_act}. The activations ($a_{pre}$) are quantized and multiplied by synaptic weights, aggregated, and then scaled to produce the discretized output activations ($a$). $a^\prime$ denotes the weighted sum of pre-activation values, and $a$ the quantized activations, with $\Delta_a$ serving as the quantization step.

\begin{equation}
    \begin{aligned}
    a^\prime &= \sum_{j} w_{j} a_{pre}, \\
    a &= \lfloor \frac{a^\prime}{\Delta_a} \rfloor \times \Delta_a
    \end{aligned}
    \label{quant_act}
\end{equation}

SNNs and QANNs' computational equivalence is captured in Eq.~\ref{qnn}, where the forward pass in QANNs is deconstructed into a summation over $T$ quantization levels, akin to SNNs' summation over $T$ time steps. 

\begin{equation}
    a = \sum_{j} w_{j} a_{pre} = \sum_{j} w_{j} \sum_{t=0}^{T} 2^{t} a_{pre}^{(t)}= \sum_{t=0}^{T} 2^{t} \sum_{j} w_{j} a_{pre}^{(t)}
    \label{qnn}
\end{equation}

In this framework, $a_{pre}^{(t)}$ represents the $t^{th}$ bit of the binary representation of $a_{pre}$, illustrating the decomposition of QANN forward computation into bit-wise operations, which closely resemble SNN's temporal spike integrations, as shown in Fig.~\ref{fig:abstract}.

The above discussion demonstrates that the Generalized Matrix Multiplication (GeMM) process of input activation values and weights in QANNs can be converted to the multiplication of binary spikes with weights, which has a similar representation to SNNs. The distinction arises in the subsequent integration stage; SNNs accumulate results over time, while QANNs do so across bit-widths. Below, we explore the hardware implementations for these integrations and their parallels.

\begin{figure}[htbp]
    \centering
    \includegraphics[width=.85\linewidth]{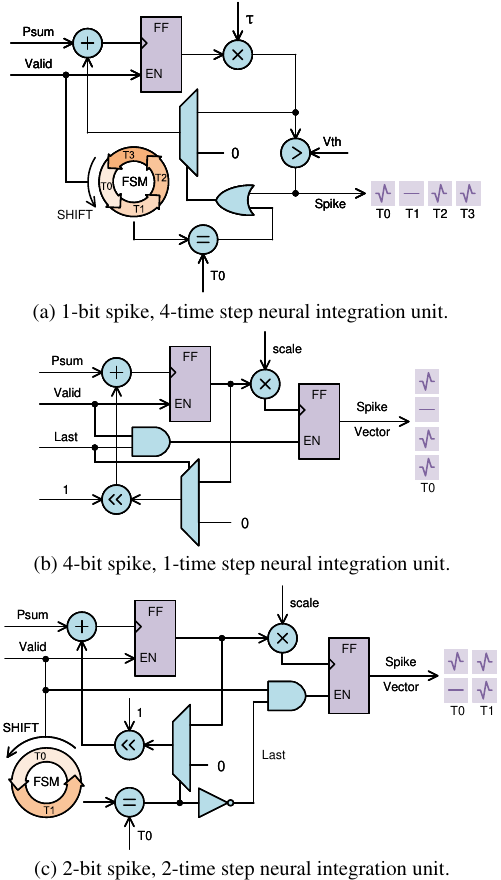}
    \vspace{-2mm}
    \caption{Comparison of different allocation of bit budgets.} 
    \label{fig:hardware} 
    \vspace{-6mm}
\end{figure}

Fig.~\ref{fig:hardware} illustrates the input integration for neurons with varying bit allocations, given a constant synaptic weight bit-width. The \texttt{P\_sum} is the weighted presynaptic input, and \texttt{Valid} signals its readiness for neuronal processing. \texttt{Last} indicates the final time step, used alongside \texttt{Valid} to capture the quantized spike vector output. A Finite State Machine (FSM) controls the timing, resetting neuron potential at the start. The subfigures (a)-(c) demonstrate the membrane potential integration for different spike-time configurations, such as 1/4, 4/1, and 2/2. This setup allows for unfolding multiple input currents for parallel processing. Hence, we assert that varying bit allocations do not significantly alter hardware implementation or complexity.

\subsection{A Generalized Framework towards QSNNs}


This subsection introduces a generalized framework for Quantitative SNNs (QSNNs). The framework aims to provide a cohesive understanding of weight quantization, time steps, spike pattern transformation, and energy consumption evaluation.


\paragraph{Bit Budget: A More Atomistic Measure}


In conventional approaches to calculating the computational overhead of SNNs, synaptic operations (SOPs) are often utilized as the fundamental unit of measurement. SOPs typically assume that synaptic weights are fixed-precision floating-point numbers and estimate energy costs based on the operations between binary spikes and continuous synaptic weights on specific neuromorphic hardware. This methodology presupposes an ideal operation of the designed networks on particular hardware. Furthermore, this method only considers the impact of spike rate and time step on energy consumption while neglecting the potential influence of weight bit-width and spike patterns. This oversight complicates the unification of SNNs and QANNs under a common framework. Therefore, we propose refining SOPs into a more fundamental unit, the Bit Budget, which we define as follows:



\vspace{-4mm}
\begin{equation}
    \operatorname{BB} = T \cdot w \cdot s
    \label{bb}
\end{equation}

\begin{figure}[htbp]
    \centering
    \includegraphics[width=0.87\linewidth]{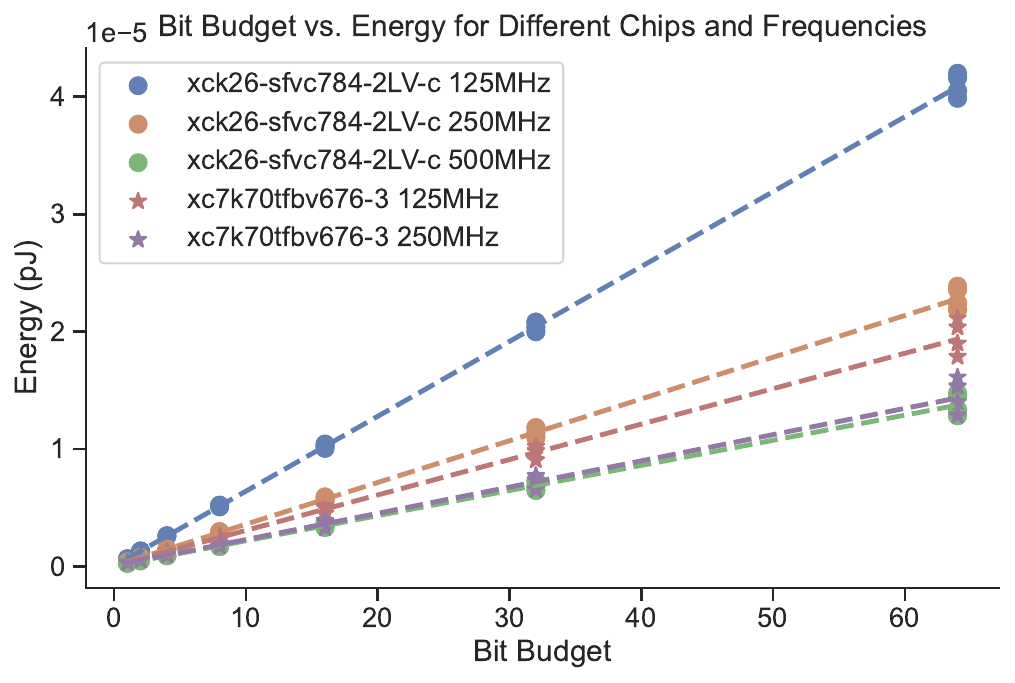}
    \vspace{-2mm}
    \caption{Bit Budget versus energy consumption for individual synaptic operations.} 
    \label{fig:bb}
    \vspace{-6mm}
\end{figure}

As depicted in Eq.~\ref{bb}, we consider three factors in tandem: the time step (T), the bit-width of the synaptic weight (W), and the bit-width of the spike (S) to estimate the computational overhead of a single synapse. This presents a more flexible degree of freedom compared to previous works, which only considered the impact of the time step and assumed fixed bit-width for synaptic weights, along with a limited binary representation for spikes. To validate the effectiveness of this metric, we used an Field Programmable Gate Arrays (FPGA) platform to analyze the bit budget corresponding to different devices and their synaptic energy consumption, as illustrated in Fig.~\ref{fig:bb}. The figure reveals a noticeable linear relationship between the bit budget and synaptic energy consumption, affirming the bit budget's validity. Next, we will revisit SNNs and optimize the computational overhead of SNNs, guided by the bit budget.

\paragraph{Quantification of Weights}


In the realm of quantized neural networks, weight quantization is pivotal for optimizing storage and computational overhead without significantly sacrificing performance. This is essential for both ANNs and SNNs, especially when pursuing efficient hardware implementations. However, despite the importance of weight quantization in SNNs, it has not received as much attention as in traditional QANNs. 

The quantization function is shown in Eq.~\ref{quant}, which transforms continuous or high-precision floating-point weights $w$ into a discrete integer set $w_q$ with reduced bit-width. This function is governed by a quantized step $\Delta$, a critical parameter that balances information fidelity against compression level. 


\vspace{-2mm}
\begin{equation}
    w_q = \lfloor \frac{w}{\Delta} \rceil   \times \Delta
    \label{quant}
\end{equation}

$\lfloor \cdot \rceil$ is the standard rounding operation. Upon quantization, each original weight $w$ is supplanted by a quantized value $w_q$, altering network dynamics. The core challenge is to turn $\Delta$ and the quantization scheme such that the quantized network's output $y_q$, remains as close as possible to that of the unquantized network $y$, despite the inherent loss of precision. The similarities between SNN and QANN architectures allow for the cross-application of quantization strategies, as outlined in Section~\ref{sec:qnn_vs_snn}. Our contribution does not delve into the complexities of specific quantization techniques; rather, it presents a unified framework for QSNNs. We utilize a symmetric quantization approach, with  $\Delta$ set to $2^{T-1}$, ensuring that the quantized weights $w_q$ fall within the range $[-1, 1]$.



\paragraph{Step-State Bit Allocation}

Our preliminary research revealed the functional equivalence between $T$-step SNNs and $T$-bit quantized state ANNs. Building upon this foundation, in the following sections, we will delve deeper into how SNNs, under a fixed bit budget, can strategically allocate bits to the representation of spike patterns and time steps to enhance overall performance.

In conventional SNNs, neuronal responses are binary, leveraging the temporal dimension to encode rich information. In contrast, QANNs encode data within quantized state bits, devoid of a distinct temporal dimension. A proposed hybrid model, therefore, seeks to synergize these approaches, judiciously distributing bits across time steps and spike patterns. For example, rather than dedicating all bits to spike patterns or temporal neuronal activity, a model might apportion its bit budget to improve temporal resolution (by increasing time steps) while using the remainder for enhanced state representation (through elevated spike patterns). This strategy promises a more refined balance between temporal granularity and state complexity, potentially unveiling new operational modes and efficiencies beyond the reach of SNNs or QANNs.

The concept of bit allocation in SNNs initially stems from the study of burst neurons~\cite{shen2023exploiting}. The burst neuron model categorizes neuronal activity into three distinct states: resting, spiking, and bursting, demonstrating its advanced capabilities. Building upon this, we have developed an enhanced model in which neurons dynamically allocate computational resources or Bit Budgets between the temporal domain and spike type. This is achieved by modulating time steps and various spike patterns, enabling neurons to emit a collection of different states rather than mere binary signals.

In the proposed model, the allocation of bits is more flexible and is not strictly confined to representing either the time steps or the quantization levels of the neuron's activity. Instead, it permits a dynamic distribution of computational resources:

\vspace{-4mm}
\begin{align}
    & \begin{aligned}
            v_{t} & = \frac{1}{\tau} v_{t-1} + \sum_{j} w_{j}s_{\text{pre}}, \quad t \in \{1, 2, ..., \lfloor \frac{T}{N} \rfloor\} \\
                  & \text{if} \ v > v_{th} \text{, then} \ v \leftarrow v_{rst}; \\
            \label{lif_budget}
      \end{aligned} \\
    & s = \lfloor \frac{v}{v_{th}} \rfloor \times v_{th}
   \label{spike_budget} 
\end{align}


As shown in Eq.~\ref{spike_budget}, the neuron's membrane potential ($v$) at each time step ($t$) is influenced by its previous state and the current synaptic inputs ($s_{\text{pre}}$), weighted by the synaptic weights ($w_j$). When the membrane potential exceeds the threshold, the neuron fires a spike, after which the neuron's membrane potential is reset. The potential quantization determines the emitted spike pattern ($s$). This quantization allows the spike to carry more information, utilizing the bit budget effectively.

This mechanism introduces an adaptable allocation strategy. Part of the bit budget is used to increase the number of time steps, enriching the temporal resolution, while the rest is employed to enhance the spike patterns via finer quantization levels. Such balanced allocation provides more nuanced control over the interplay between time-sensitive dynamics and the richness of state information, paving the way for more efficient and adaptable neural computations in QSNNs.

\paragraph{Computational Effort Estimates for SNNs}


In light of these limitations and inspired by the computational paradigm in~\cite{zhang2022pokebnn}, we propose two advanced metrics aimed at a more holistic and hardware-independent evaluation of computational expenses based on the bit budget: the Synaptic Arithmetic Computation Effort (S-ACE) and the Neuromorphic Synaptic Arithmetic Computation Effort (NS-ACE). These metrics are specifically tailored to assess the resource expenditure of SNNs in both generic and neuromorphic computing environments, offering broader applicability.

The S-ACE is calculated with Eq.~\ref{sace}:

\vspace{-2mm}
\begin{equation}
\operatorname{S-ACE} = \sum_{\substack{w \in W, \ s \in S}} n_{w, s} \cdot \operatorname{BB}
\label{sace}
\end{equation}

In Eq.~\ref{sace}, $n_{w, s}$ denotes the count of multiply-accumulate operations (MACs) for an $w$-bit number and a $s$-bit number extracted directly from the neural network's architecture. $W$ and $S$ constitute the sets containing all bit-widths engaged during the neural network's inference stage.

On the contrary, NS-ACE integrates the neuron's firing rate ($fr_s$) into its computations to more accurately reflect the energy dynamics inherent to neuromorphic hardware systems. This model is predicated on the principle that energy consumption occurs exclusively during spike events. The formulation for NS-ACE is as follows:

\vspace{-2mm}
\begin{equation}
\operatorname{NS-ACE} = \sum_{\substack{w \in W, \ s \in S}} fr_s \cdot n_{w, s} \cdot  \operatorname{BB} 
\label{nsace}
\end{equation}

In Eq.~\ref{nsace}, the term $fr_s$ represents the firing rate of neurons, emphasizing the concept that in neuromorphic systems, the absence of spiking activity is associated with negligible energy expenditure.

Our introduction of S-ACE and NS-ACE marks a significant stride toward a more refined, realistic, and hardware-agnostic evaluation of computational costs in SNNs. By integrating the complexities inherent in real-world applications, these metrics offer a versatile and comprehensive framework for researchers and engineers working in the SNN landscape.
\section{Experiments}

To assess the performance of QSNNs and determine their comparative effectiveness against conventional neural network models across various scenarios, we conducted a comprehensive evaluation using a range of datasets.  Our analysis included traditional static datasets such as ImageNet~\cite{deng2009imagenet} and CIFAR10/100~\cite{krizhevsky2009learning}, along with a collection of neuromorphic datasets like CIFAR10-DVS~\cite{li2017cifar10}, DVS128 Gesture~\cite{amir2017low}, and N-Caltech101~\cite{orchard2015converting}. For detailed information about these datasets and the experimental settings, please refer to the Appendix~\ref{suppl:data}.

Our experimental analysis centers on the spike patterns and weight quantization of SNNs, aiming to dissect the subtle differences and interconnections between SNNs and ANNs. It is worth mentioning that our research findings can be integrated with contemporary neural network quantization techniques to further enhance model performance. However, while we have achieved state-of-the-art results in several SNN benchmarks, the focus of our study is not to establish new performance benchmarks for SNNs. Instead, our goal is to enrich the understanding of how SNNs function under quantization constraints and their relationship with their non-spiking counterparts. 

\subsection{Evaluation on Static Datasets} 

\begin{table*}[htbp]
    \caption{Evaluation on ImageNet. In quantization, 'W' represents the bit-width of weights, 'S' denotes the bit-width of spike patterns, and 'T' represents the time steps. SOPs refer to synaptic operations. 
    When calculating S-ACE / NS-ACE for FP32 operations, we assume that these operations can be performed in BF16 without a loss of precision. $\uparrow$ means the higher the better, $\downarrow$ means the lower the better. }
    \vspace{-4mm}
    \begin{center}
        \resizebox{.95\textwidth}{!}{
        \begin{tabular}{ccccccccc}
            \toprule
            Methods                                                   & Architecture        &    \tabincell{c}{Bit ($\downarrow$) \\ Budget }   & \tabincell{c}{Quantization \\ W / S / T} &  \tabincell{c}{Params \\ (M) ($\downarrow$)}  & \tabincell{c}{SOPs \\ (G) ($\downarrow$)} & \tabincell{c}{S-ACE \\ (G) ($\downarrow$)} & \tabincell{c}{NS-ACE \\ (G) ($\downarrow$)}  & Acc (\%) ($\uparrow$) \\
            \midrule
            Hybrid training~\cite{rathi2020enabling}       & ResNet-34         &     4000     & 16 / 1 / 250     & 21.79   & 850.6  & 13608   &   -       & 61.48          \\
            TET~\cite{deng2022temporal}              & SEW-ResNet-34        &   64    & 16 / 1 / 4       & 21.79   & 13.61  & 217.72  &       -     & 68.00          \\
            Spiking ResNet~\cite{hu2021spiking}           & ResNet-50      &   5600          & 16 / 1 / 350     & 25.56   & 1184   & 18952   &      -      & 72.75          \\
            tdBN~\cite{zheng2021going}           & Spiking-ResNet-34      &   96  & 16 / 1 / 6       & 21.79   & 20.42  & 326.60  &      -       & 63.72          \\
            \midrule
            \multirow{2}{*}{SEW ResNet \cite{fang2021deep}}
                                                                      & SEW-ResNet-34       &  64      & 16 / 1 / 4       & 21.79  & 13.61  & 217.72  &    63.14         & 67.04          \\
                                                                      & SEW-ResNet-50       &  64      & 16 / 1 / 4       & 25.56  & 13.48  & 215.62  &    69.03         & 67.78          \\
            \hline
            \multirow{4}{*}{\tabincell{c}{Quantized \\ SEW ResNet}}                                                         
                                                                      & SEW-ResNet-34        &  1     & 1 / 1 / 1        & 1.36   & 3.40   & 3.40    &    1.01         &    52.17       \\
                                                                      \cline{2-9}
                                                                      & SEW-ResNet-34        &  4     & 2 / 1 / 2        & 2.72   & 6.80   & 13.6    &     4.11        &     60.15      \\
                                                                      & SEW-ResNet-34        &  4     & 2 / 2 / 1        & 2.72   & 6.80   & 13.6    &     3.65        &     62.36       \\
                                                                      \cline{2-9}
                                                                      & SEW-ResNet-34        &  64     & 8 / 8 / 1        & 10.89  & 27.1   & 217     &    56.42       &  \textbf{70.13}          \\
            \midrule
            \multirow{5}{*}{Spikformer~\cite{zhou2022spikformer}}
                                                                      & Spikformer-8-384  & 64 & 16 / 1 / 4       & 16.81   & 18.70   & 299.6   &  80.89            & 70.24          \\
                                                                      & Spikformer-6-512  & 64 & 16 / 1 / 4       & 23.37   & 28.16   & 451.2   &  121.03           & 72.46          \\
                                                                      & Spikformer-8-512  & 64 & 16 / 1 / 4       & 29.68   & 33.24   & 516.4   &  159.16           & 73.38          \\
                                                                      & Spikformer-10-512 & 64 & 16 / 1 / 4       & 36.01   & 38.31   & 613.6   &  165.30           & 73.68          \\
                                                                      & Spikformer-8-768  & 64 & 16 / 1 / 4       & 66.34   & 74.78   & 1197    &  319.64           & 74.81          \\
            \hline
            \multirow{6}{*}{\tabincell{c}{Quantized \\ Spikformer}}                                                          
                                                                      & Spikformer-8-512  & 1 & 1 / 1 / 1        & 1.86   & 6.67   & 6.79    &      2.12         &  54.54           \\
                                                                      \cline{2-9}
                                                                      & Spikformer-8-512  & 4 & 2 / 2 / 1        & 3.71   & 13.19  & 26.52   &      7.65         &  63.16                \\
                                                                      & Spikformer-8-512  & 4 & 2 / 1 / 2        & 3.71   & 13.33  & 26.92   &      7.86         &  61.37                \\
                                                                      \cline{2-9}
                                                                      & Spikformer-8-512  & 8 & 4 / 2 / 1        & 3.71   & 13.19  & 52.90   &      14.36         &  70.87            \\
                                                                      & Spikformer-8-512  & 8 & 2 / 4 / 1        & 7.42   & 26.25  & 52.90   &      14.17         &  64.75            \\
                                                                      \cline{2-9}
                                                                      & Spikformer-8-512  & 32 & 4 / 8 / 1        & 14.84  & 26.25  & 210.1   &      54.76         &  76.83            \\
                                                                      & Spikformer-8-512  & 32 & 8 / 4 / 1        & 14.84  & 26.25  & 210.1   &      52.10         &  \textbf{79.37}            \\
            \hline
        \end{tabular}
        }
    \end{center}
    \label{table:imagenet_result}
    \vspace{-4mm}
\end{table*}
As seen in Tab.~\ref{table:imagenet_result}, we examine the influence of bit allocation, model size, and accuracy on various SNN structures, including convolutional and Transformer variants on the ImageNet dataset. Our findings indicate that for static images, allocating bits to enhance spike patterns, rather than increasing time steps, leads to superior performance due to the static nature of these images. This approach, favoring dense spike patterns, proves more efficient in information transmission while maintaining a consistent computational budget, as quantified by the S-ACE and NS-ACE metrics. Additionally, Tab.~\ref{table:imagenet_result} reveals that reducing weight bit-width results in a reduction of parameters and variations in performance.

\begin{figure}[htbp]
    \centering
    \includegraphics[width=\linewidth]{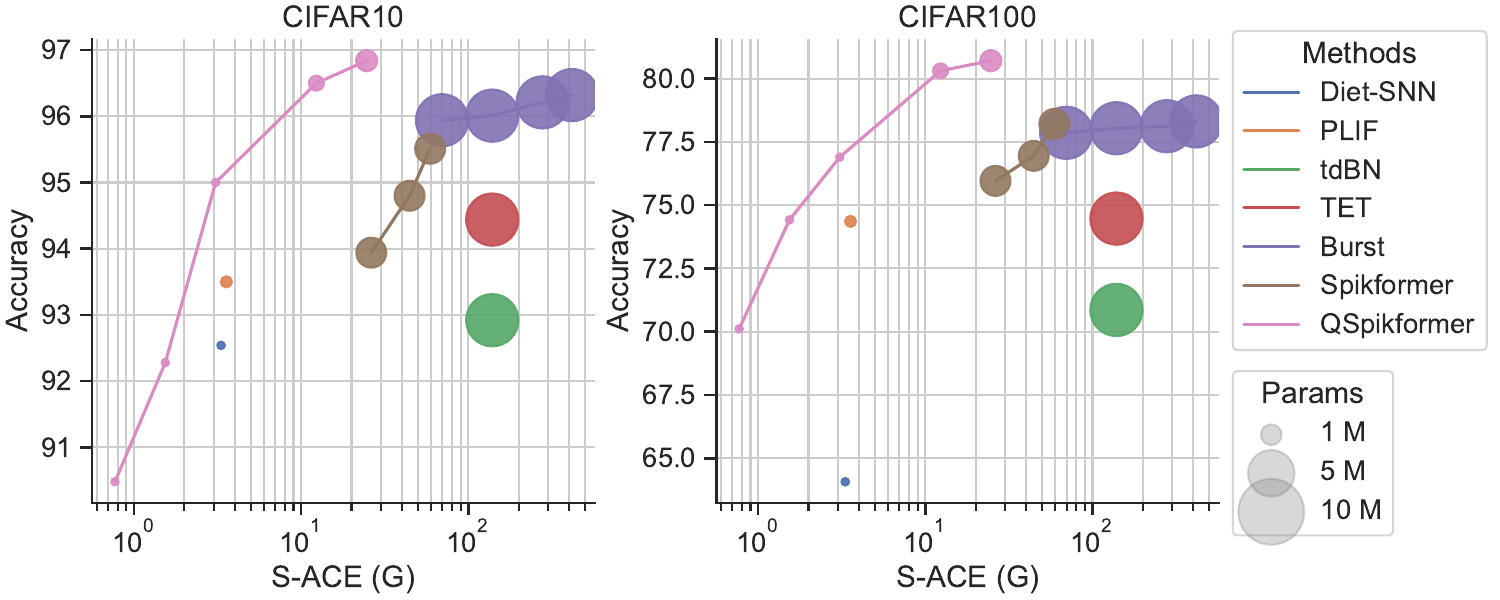}
    \caption{Accuracy, S-ACE and model parameters on the CIFAR dataset. S-ACE is expressed in logarithmic coordinates.}
    \label{fig:cifar}
    \vspace{-4mm}
\end{figure}

The results in Tab.~\ref{table:cifar_result} have shown the effects of allocating bit budgets between spike patterns and time steps on CIFAR10/100 datasets while keeping weight and bit-widths consistent. The optimal model performance on CIFAR10/100 is achieved by limiting time steps to 1 or 2 and maximizing the spike pattern bit-width, which is consistent with the results on the ImageNet dataset. A detailed exposition is available in Appendix~\ref{suppl:cifar}. Furthermore, Fig.~\ref{fig:cifar} graphically represents the relationships among accuracy, S-ACE (logarithmically scaled), and model parameters, illustrating the unique balances each method strikes between accuracy, computational cost, and model size. Our methodology demonstrates a proficient equilibrium of these pivotal factors. 

\begin{table}[b]
    \caption{Evaluation on CIFAR10/100.} 
    \vspace{-6mm}
    \begin{center}
        \resizebox{.95\columnwidth}{!}{
            \begin{tabular}{ccccc}
                \toprule
                \tabincell{c}{Allocation \\ W/S/T} &  \tabincell{c}{Params  \\ (M) ($\downarrow$)} & \tabincell{c}{S-ACE \\ (G) ($\downarrow$)} & \tabincell{c}{CIFAR10 \\ Acc($\uparrow$)  }  & \tabincell{c}{CIFAR100 \\Acc ($\uparrow$) }\\
                \midrule
                16 / 1 / 4   & 4.15  & 59.10 & 95.51 & 78.21  \\ 
                \hline 
                        
                1 / 1 / 1    & 0.26  & 0.77 & 90.48 & 70.11  \\          
                \cline{1-5}      
                1 / 4 / 1    & 0.26  & 3.08 & 95.00 & \textbf{76.90}  \\     
                1 / 2 / 2    & 0.52  & 3.08 & 94.43 & 75.91  \\   
                1 / 1 / 4    & 0.26  & 3.69 & 93.91 & 74.13  \\    
                \cline{1-5}      
                2 / 2 / 1    & 0.52  & 3.08 & 95.41 & \textbf{76.67}  \\     
                2 / 1 / 2    & 0.52  & 3.09 & 93.56 & 75.91  \\    
                \cline{1-5}                   
                4 / 1 / 1    & 1.04  & 3.09 & 94.51 & 74.61  \\            
                4 / 4 / 1    & 1.04  & 12.32 & 96.84 & 80.13  \\      
                4 / 2 / 2    & 1.04  & 12.33 & 96.50 & \textbf{80.71}  \\      
                4 / 1 / 4    & 1.04  & 14.77 & 95.94 & 78.77  \\      
                \cline{1-5}      
                8 / 1 / 2    & 2.08  & 12.35 & 95.55 & 77.72  \\       
                8 / 2 / 1    & 2.08  & 12.33 & 96.29 & \textbf{80.00}  \\        
                \hline   
        \end{tabular}
        }
    \end{center}
    \label{table:cifar_result} 
    \vspace{-6mm}
\end{table}

\subsection{Evaluation on Neuromorphic Datasets}



For neuromorphic datasets, we utilize the model outlined in Tab.~\ref{table:cifar_result} and undertake a comprehensive analysis to evaluate the influence of bit distribution between spike patterns and time steps on model performance, maintaining a uniform weight bit-width. For the sake of simplicity, we set the weight bit-width to $1$.

Tab.~\ref{table:dvs_result} reveals that adjusting the bit allocation between spike patterns and time steps results in considerable performance trade-offs. However, the effect of weight bit-width on model performance appears less substantial, possibly due to the constrained size of the neuromorphic datasets, where weight bit-width may not be a critical performance factor.

\begin{table}[b]
    \vspace{-4mm}
    \centering
    \caption{Evaluation of different Bit Budgets on neuromorphic datasets.}
    \resizebox{.97\linewidth}{!}{%
    \begin{tabular}{c|ccc}
    \hline
    W / S / T & DVSC10~\cite{li2017cifar10} & DVSG~\cite{amir2017low} &  NCAL~\cite{orchard2015converting} \\
    \hline
    16 / 1 /16 & 80.7 & 98.3 &  80.23 \\
    \hline 
    1 / 1 / 16 & \textbf{79.8} (+0.00) & 96.67 (+0.00) & 79.77 (+0.00) \\
    1 / 2 / 8  & 79.3 (-0.50) & \textbf{98.48} (+1.81) & 80.00 (+0.23) \\
    1 / 4 / 4  & 63.1 (-16.7) & 97.35 (+0.68) & \textbf{82.64} (+2.87)\\
    1 / 8 / 2  & 43.0 (-36.8) & 96.59 (-0.08) & 80.46 (+0.69)\\
    1 / 16 / 1 & 35.8 (-44.0) & 95.45 (-1.22) & 77.47 (-2.30)\\
    \hline
    \end{tabular}%
    }
    \label{table:dvs_result}
\end{table}

\subsection{Bit Budget Allocation}

The performance of SNN models is contingent upon the optimized interplay of weight bit-width (W), spike patterns (S), and the count of time steps (T). We conceptualize the cumulative product of these parameters as the Bit Budget, corresponding to the S-ACE for a single operation. The pivotal aspect of our research involves determining the optimal allocation of these variables within the constraints of a predetermined Bit Budget to augment the comprehensive performance of the model.

\begin{figure}[htbp]
    \centering
    \includegraphics[width=\linewidth]{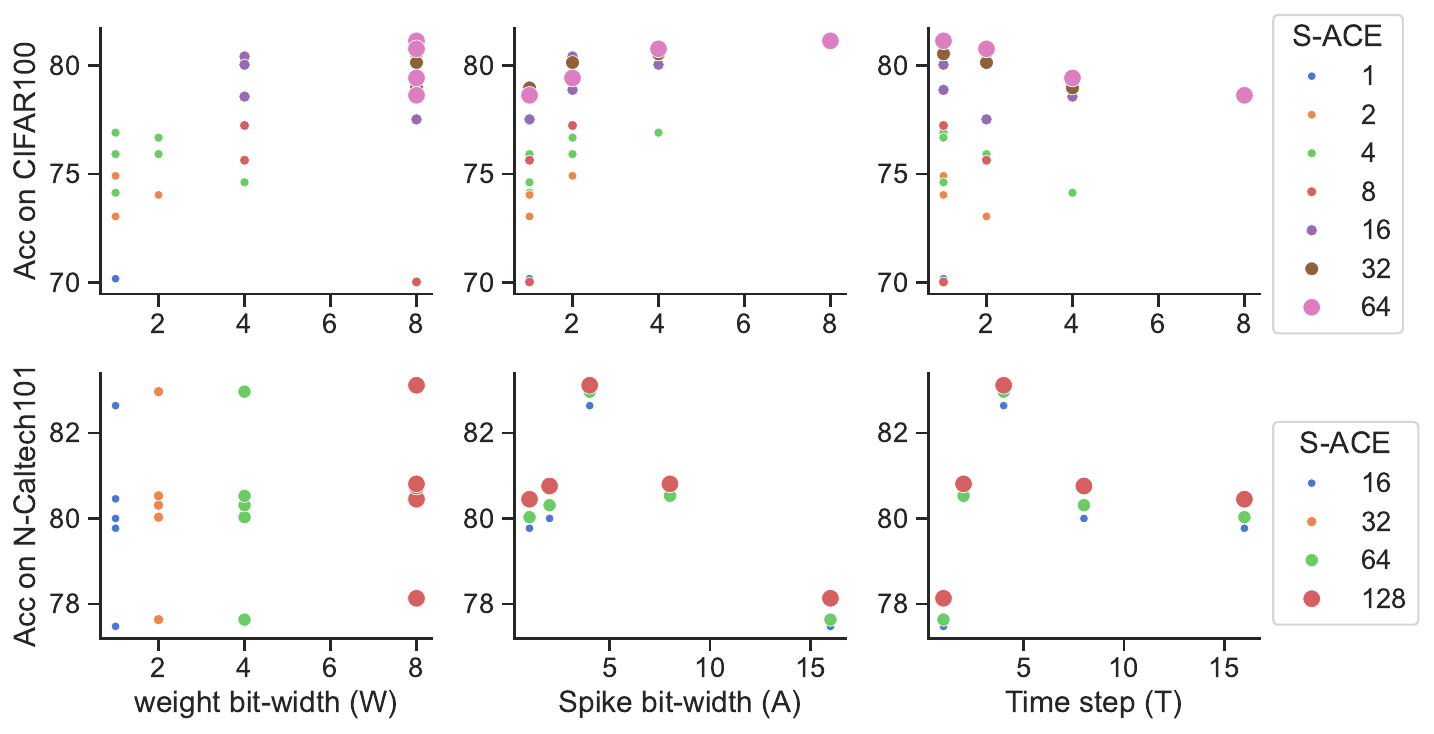} 
    \caption{Relationship between different bit budget allocations and model performance.}
    \label{fig:bit_budget}
    \vspace{-4mm}
\end{figure}

As shown in Fig.~\ref{fig:bit_budget}, our trials on CIFAR100 and N-Caltech101 unveiled a clear pattern: boosting spike bit-width enhanced performance for static images, while increasing time steps did not. This aligns with the static nature of such images where temporal aspects are less significant. For neuromorphic data, a careful balance between spike bit-widths and time steps was vital. Setting time steps to $T=4$ and judiciously allocating the remaining bit budget yielded positive results under limited model parameters and computational resources. A marked increase in model accuracy was observed with higher weight bit-widths, plateauing after a certain point. Interestingly, models with low-weight bit-widths, like 1-bit, still performed well, a fact often overlooked in prior research. Properly adjusting weight bit-width can significantly enhance inference efficiency and compact the model size, which is crucial for the inherent efficiency of SNNs.

\subsection{Hardware Implementation}
\label{sec:hardware_implementation}


\begin{table}[htbp]
    \centering
    \caption{Hardware Implementation Results on CIFAR10 and DVS-CIFAR10}
    \resizebox{.95\columnwidth}{!}{
        \begin{tabular}{ccccc}
        \hline
        \multicolumn{5}{c}{CIFAR10} \\ 
        \hline
        Device & Bit Budget & W / S / T & fps &Accuracy (\%) \\
        \hline
        \multirow{7}{*}{xck26} & \multirow{3}{*}{32} & 8 / 1 / 4 & 342.35  & 95.45 \\
                               &                     & 8 / 2 / 2 & 343.52  & 95.84 \\
                               &                     & 8 / 4 / 1 & 343.29  & 96.41 \\
                               \cline{2-5}
                               & \multirow{4}{*}{64} & 8 / 1 / 8 & 155.30  & 95.47 \\
                               &                     & 8 / 2 / 3 & 155.55  & 95.79 \\
                               &                     & 8 / 4 / 2 & 155.47  & 96.43 \\
                               &                     & 8 / 8 / 1 & 155.23  & 96.51 \\
        \hline
        \multicolumn{5}{c}{DVS-CIFAR10} \\ 
        \hline
        Device & Bit Budget & W / S / T & Latency &Accuracy (\%) \\
        \hline
        \multirow{4}{*}{xck26} & \multirow{4}{*}{128} & 8 / 1 / 16 & 49.58  & 80.30 \\
                               &                      & 8 / 2 / 8 & 49.56  & 80.10 \\
                               &                      & 8 / 4 / 4 & 49.58  & 69.13 \\
                               &                      & 8 / 8 / 2 & 49.57  & 51.31 \\
        \hline
        \end{tabular}
    }
    \label{tab:hardware_implementation}
    \vspace{-4mm}
\end{table}




We have investigated various bit allocation strategies on FPGA platforms for efficient hardware implementations of neural networks. We examine the balance between spike pattern complexity, time steps, fps, and accuracy, constrained by $8$-bit weight precision. As indicated in Tab.~\ref{tab:hardware_implementation}, the congruence between our FPGA-based experimental results and software simulation results underscores the robustness and universality of our methodologies across varied hardware environments. For detailed hardware experimental setup and more experimental results, please refer to the Appendix~\ref{suppl:hardware}.

\begin{figure}[htbp]
    \centering
    \includegraphics[width=.95\linewidth]{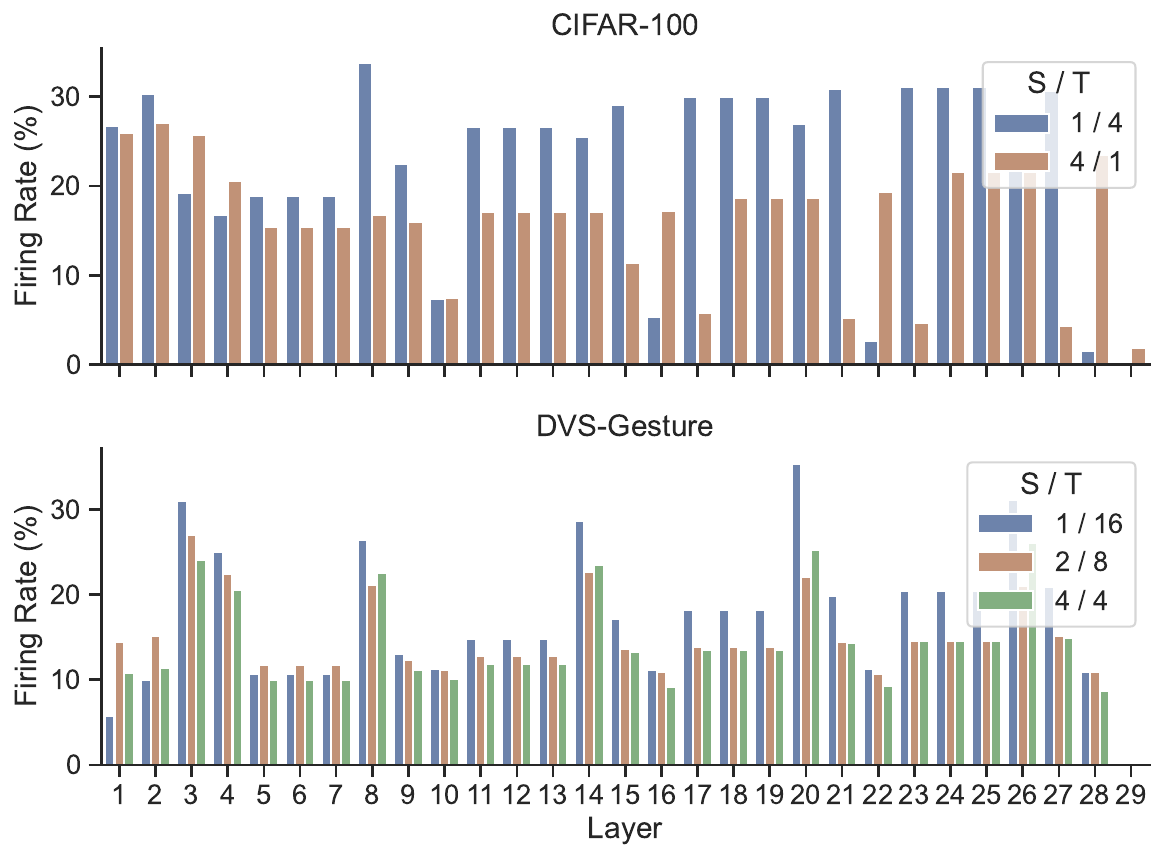}
    \vspace{-6mm}
    \caption{Comparison of firing rates of different layers corresponding to different bit allocation strategies between spike patterns and time steps.}
    \label{fig:firing_rate}
    \vspace{-6mm}
\end{figure}

\subsection{Visualization of Sparsity in SNNs}

In neural networks, sparsity, characterized by inactive neurons and synapses, contributes to energy efficiency by reducing computational and memory requirements. This subsection explores the influence of various bit budget allocations on neuron sparsity within SNNs. Employing the CIFAR100 and DVS-Gesture datasets and utilizing single-bit weight models, we analyzed neuron firing rates across different allocations of bits to spike patterns and time steps. As depicted in Fig.~\ref{fig:firing_rate}, the results reveal that neuron sparsity was not notably prominent with single-bit spike patterns. Conversely, increasing bits for spike patterns while reducing them for time steps led to lower firing rates, indicating an increase in network sparsity.

The results indicate that allocating more bits to spike patterns than time steps can lead to more energy-efficient SNN operations, potentially without performance loss.  This efficiency likely arises from a more effective use of the network's representational capabilities and decreased computational waste. Our findings highlight that SNN sparsity is not a fixed attribute but can be effectively managed through design, with the Bit Budget concept being key in optimizing both sparsity and performance. This observation calls into question the prevailing assumptions regarding intrinsic sparsity in SNNs and propels novel strategies for architecting efficient neural networks.


\section{Conclusion}
\label{sec:discussion}

The current trend in the design of deep SNNs favors simplified neuron models that aim to achieve a scale comparable to modern artificial neural networks. While these SNNs exhibit notable performance in specific contexts, they often struggle to surpass the performance benchmarks set by traditional ANNs. A critical insight from our research is that SNNs designed following the ANN paradigm inherently mirror the core characteristics of ANNs, akin to ANNs that employ activation value quantization. This revelation highlights the challenges faced by such design philosophies. Our in-depth analysis of SNNs accentuates the pivotal role of bit allocation strategies in regulating network sparsity and enhancing energy efficiency. We discern that sparsity in SNNs is not intrinsic but can be effectively modulated and optimized through meticulous design strategies. 

Consequently, we introduce the Bit Budget concept as a systematic approach to enhance network representational capabilities while minimizing computational waste, thus balancing sparsity and performance. In our bit budgeting, prioritizing the state representation of SNNs over time steps has been shown to augment energy efficiency without compromising performance. Through simulations and FPGA experiments, we have validated the robustness and applicability of our approach across various computational settings. This research deepens the understanding of SNN dynamics and provides practical guidelines for building more efficient SNNs. It carves out new pathways for advancing SNNs, especially in developing energy-efficient systems for real-world applications.  

Our findings challenge the prevailing notion that contemporary SNNs are intrinsically high-efficiency systems. We advocate for a paradigm that aligns harmoniously with neuroscience rather than pursuing efficiency as an end in itself. We aspire to see the field pivot towards a more integrated approach with neuroscience, fostering intelligent networks that are energy-efficient and deeply rooted in the principles of brain-like computation.
{
    \small
    \bibliographystyle{ieeenat_fullname}
    \bibliography{refs}

\begin{thebibliography}{43}
\providecommand{\natexlab}[1]{#1}
\providecommand{\url}[1]{\texttt{#1}}
\expandafter\ifx\csname urlstyle\endcsname\relax
  \providecommand{\doi}[1]{doi: #1}\else
  \providecommand{\doi}{doi: \begingroup \urlstyle{rm}\Url}\fi

\bibitem[Amir et~al.(2017)Amir, Taba, Berg, Melano, McKinstry, Di~Nolfo, Nayak,
  Andreopoulos, Garreau, Mendoza, et~al.]{amir2017low}
Arnon Amir, Brian Taba, David Berg, Timothy Melano, Jeffrey McKinstry, Carmelo
  Di~Nolfo, Tapan Nayak, Alexander Andreopoulos, Guillaume Garreau, Marcela
  Mendoza, et~al.
\newblock A low power, fully event-based gesture recognition system.
\newblock In \emph{Proceedings of the IEEE conference on computer vision and
  pattern recognition}, pages 7243--7252, 2017.

\bibitem[Bohte(2011)]{bohte2011error}
Sander~M Bohte.
\newblock Error-backpropagation in networks of fractionally predictive spiking
  neurons.
\newblock In \emph{International conference on artificial neural networks},
  pages 60--68. Springer, 2011.

\bibitem[Courbariaux et~al.(2016)Courbariaux, Hubara, Soudry, El-Yaniv, and
  Bengio]{courbariaux2016binarized}
Matthieu Courbariaux, Itay Hubara, Daniel Soudry, Ran El-Yaniv, and Yoshua
  Bengio.
\newblock Binarized neural networks: Training deep neural networks with weights
  and activations constrained to+ 1 or-1.
\newblock \emph{arXiv preprint arXiv:1602.02830}, 2016.

\bibitem[Davies et~al.(2018)Davies, Srinivasa, Lin, Chinya, Cao, Choday, Dimou,
  Joshi, Imam, Jain, et~al.]{davies2018loihi}
Mike Davies, Narayan Srinivasa, Tsung-Han Lin, Gautham Chinya, Yongqiang Cao,
  Sri~Harsha Choday, Georgios Dimou, Prasad Joshi, Nabil Imam, Shweta Jain,
  et~al.
\newblock Loihi: A neuromorphic manycore processor with on-chip learning.
\newblock \emph{Ieee Micro}, 38\penalty0 (1):\penalty0 82--99, 2018.

\bibitem[Deng et~al.(2009)Deng, Dong, Socher, Li, Li, and
  Fei-Fei]{deng2009imagenet}
Jia Deng, Wei Dong, Richard Socher, Li-Jia Li, Kai Li, and Li Fei-Fei.
\newblock Imagenet: A large-scale hierarchical image database.
\newblock In \emph{2009 IEEE conference on computer vision and pattern
  recognition}, pages 248--255. Ieee, 2009.

\bibitem[Deng et~al.(2022)Deng, Li, Zhang, and Gu]{deng2022temporal}
Shikuang Deng, Yuhang Li, Shanghang Zhang, and Shi Gu.
\newblock Temporal efficient training of spiking neural network via gradient
  re-weighting.
\newblock \emph{arXiv preprint arXiv:2202.11946}, 2022.

\bibitem[Fang et~al.(2021{\natexlab{a}})Fang, Yu, Chen, Huang, Masquelier, and
  Tian]{fang2021deep}
Wei Fang, Zhaofei Yu, Yanqi Chen, Tiejun Huang, Timoth{\'e}e Masquelier, and
  Yonghong Tian.
\newblock Deep residual learning in spiking neural networks.
\newblock \emph{Advances in Neural Information Processing Systems},
  34:\penalty0 21056--21069, 2021{\natexlab{a}}.

\bibitem[Fang et~al.(2021{\natexlab{b}})Fang, Yu, Chen, Masquelier, Huang, and
  Tian]{fang2021incorporating}
Wei Fang, Zhaofei Yu, Yanqi Chen, Timoth{\'e}e Masquelier, Tiejun Huang, and
  Yonghong Tian.
\newblock Incorporating learnable membrane time constant to enhance learning of
  spiking neural networks.
\newblock In \emph{Proceedings of the IEEE/CVF international conference on
  computer vision}, pages 2661--2671, 2021{\natexlab{b}}.

\bibitem[Gordon et~al.(2020)Gordon, Duh, and Andrews]{gordon2020compressing}
Mitchell~A Gordon, Kevin Duh, and Nicholas Andrews.
\newblock Compressing bert: Studying the effects of weight pruning on transfer
  learning.
\newblock \emph{arXiv preprint arXiv:2002.08307}, 2020.

\bibitem[Han et~al.(2023)Han, Zhao, Zeng, Pan, and Shen]{han2023enhancing}
Bing Han, Feifei Zhao, Yi Zeng, Wenxuan Pan, and Guobin Shen.
\newblock Enhancing efficient continual learning with dynamic structure
  development of spiking neural networks.
\newblock \emph{arXiv preprint arXiv:2308.04749}, 2023.

\bibitem[Han et~al.(2015{\natexlab{a}})Han, Mao, and Dally]{han2015deep}
Song Han, Huizi Mao, and William~J Dally.
\newblock Deep compression: Compressing deep neural networks with pruning,
  trained quantization and huffman coding.
\newblock \emph{arXiv preprint arXiv:1510.00149}, 2015{\natexlab{a}}.

\bibitem[Han et~al.(2015{\natexlab{b}})Han, Pool, Tran, and
  Dally]{han2015learning}
Song Han, Jeff Pool, John Tran, and William Dally.
\newblock Learning both weights and connections for efficient neural network.
\newblock \emph{Advances in neural information processing systems}, 28,
  2015{\natexlab{b}}.

\bibitem[Hu et~al.(2021)Hu, Tang, and Pan]{hu2021spiking}
Yangfan Hu, Huajin Tang, and Gang Pan.
\newblock Spiking deep residual networks.
\newblock \emph{IEEE Transactions on Neural Networks and Learning Systems},
  2021.

\bibitem[Krizhevsky et~al.(2009)Krizhevsky, Hinton,
  et~al.]{krizhevsky2009learning}
Alex Krizhevsky, Geoffrey Hinton, et~al.
\newblock Learning multiple layers of features from tiny images.
\newblock 2009.

\bibitem[Li et~al.(2016)Li, Kadav, Durdanovic, Samet, and Graf]{li2016pruning}
Hao Li, Asim Kadav, Igor Durdanovic, Hanan Samet, and Hans~Peter Graf.
\newblock Pruning filters for efficient convnets.
\newblock \emph{arXiv preprint arXiv:1608.08710}, 2016.

\bibitem[Li et~al.(2017)Li, Liu, Ji, Li, and Shi]{li2017cifar10}
Hongmin Li, Hanchao Liu, Xiangyang Ji, Guoqi Li, and Luping Shi.
\newblock Cifar10-dvs: an event-stream dataset for object classification.
\newblock \emph{Frontiers in neuroscience}, 11:\penalty0 309, 2017.

\bibitem[Li et~al.(2023)Li, Shen, Zhao, Zhang, and Zeng]{li2023firefly}
Jindong Li, Guobin Shen, Dongcheng Zhao, Qian Zhang, and Yi Zeng.
\newblock Firefly: A high-throughput hardware accelerator for spiking neural
  networks with efficient dsp and memory optimization.
\newblock \emph{IEEE Transactions on Very Large Scale Integration (VLSI)
  Systems}, 2023.

\bibitem[Li et~al.(2022)Li, Xu, Zhang, Cao, Gao, and Guo]{li2022q}
Yanjing Li, Sheng Xu, Baochang Zhang, Xianbin Cao, Peng Gao, and Guodong Guo.
\newblock Q-vit: Accurate and fully quantized low-bit vision transformer.
\newblock \emph{Advances in Neural Information Processing Systems},
  35:\penalty0 34451--34463, 2022.

\bibitem[Lin et~al.(2021)Lin, Zhang, Sun, Li, and Zhou]{lin2021fq}
Yang Lin, Tianyu Zhang, Peiqin Sun, Zheng Li, and Shuchang Zhou.
\newblock Fq-vit: Post-training quantization for fully quantized vision
  transformer.
\newblock \emph{arXiv preprint arXiv:2111.13824}, 2021.

\bibitem[Liu et~al.(2018)Liu, Sun, Zhou, Huang, and Darrell]{liu2018rethinking}
Zhuang Liu, Mingjie Sun, Tinghui Zhou, Gao Huang, and Trevor Darrell.
\newblock Rethinking the value of network pruning.
\newblock \emph{arXiv preprint arXiv:1810.05270}, 2018.

\bibitem[Liu et~al.(2021)Liu, Wang, Han, Zhang, Ma, and Gao]{liu2021post}
Zhenhua Liu, Yunhe Wang, Kai Han, Wei Zhang, Siwei Ma, and Wen Gao.
\newblock Post-training quantization for vision transformer.
\newblock \emph{Advances in Neural Information Processing Systems},
  34:\penalty0 28092--28103, 2021.

\bibitem[Loshchilov and Hutter(2017)]{loshchilov2017decoupled}
Ilya Loshchilov and Frank Hutter.
\newblock Decoupled weight decay regularization.
\newblock \emph{arXiv preprint arXiv:1711.05101}, 2017.

\bibitem[Maass(1997)]{maass1997networks}
Wolfgang Maass.
\newblock Networks of spiking neurons: the third generation of neural network
  models.
\newblock \emph{Neural networks}, 10\penalty0 (9):\penalty0 1659--1671, 1997.

\bibitem[Merolla et~al.(2014)Merolla, Arthur, Alvarez-Icaza, Cassidy, Sawada,
  Akopyan, Jackson, Imam, Guo, Nakamura, et~al.]{merolla2014million}
Paul~A Merolla, John~V Arthur, Rodrigo Alvarez-Icaza, Andrew~S Cassidy, Jun
  Sawada, Filipp Akopyan, Bryan~L Jackson, Nabil Imam, Chen Guo, Yutaka
  Nakamura, et~al.
\newblock A million spiking-neuron integrated circuit with a scalable
  communication network and interface.
\newblock \emph{Science}, 345\penalty0 (6197):\penalty0 668--673, 2014.

\bibitem[Nagel et~al.(2020)Nagel, Amjad, Van~Baalen, Louizos, and
  Blankevoort]{nagel2020up}
Markus Nagel, Rana~Ali Amjad, Mart Van~Baalen, Christos Louizos, and Tijmen
  Blankevoort.
\newblock Up or down? adaptive rounding for post-training quantization.
\newblock In \emph{International Conference on Machine Learning}, pages
  7197--7206. PMLR, 2020.

\bibitem[Orchard et~al.(2015)Orchard, Jayawant, Cohen, and
  Thakor]{orchard2015converting}
Garrick Orchard, Ajinkya Jayawant, Gregory~K Cohen, and Nitish Thakor.
\newblock Converting static image datasets to spiking neuromorphic datasets
  using saccades.
\newblock \emph{Frontiers in neuroscience}, 9:\penalty0 437, 2015.

\bibitem[Rathi and Roy(2020)]{rathi2020diet}
Nitin Rathi and Kaushik Roy.
\newblock Diet-snn: Direct input encoding with leakage and threshold
  optimization in deep spiking neural networks.
\newblock \emph{arXiv preprint arXiv:2008.03658}, 2020.

\bibitem[Rathi et~al.(2020)Rathi, Srinivasan, Panda, and
  Roy]{rathi2020enabling}
Nitin Rathi, Gopalakrishnan Srinivasan, Priyadarshini Panda, and Kaushik Roy.
\newblock Enabling deep spiking neural networks with hybrid conversion and
  spike timing dependent backpropagation.
\newblock \emph{arXiv preprint arXiv:2005.01807}, 2020.

\bibitem[Sanh et~al.(2020)Sanh, Wolf, and Rush]{sanh2020movement}
Victor Sanh, Thomas Wolf, and Alexander Rush.
\newblock Movement pruning: Adaptive sparsity by fine-tuning.
\newblock \emph{Advances in Neural Information Processing Systems},
  33:\penalty0 20378--20389, 2020.

\bibitem[Shen et~al.(2022)Shen, Zhao, and Zeng]{shen2022backpropagation}
Guobin Shen, Dongcheng Zhao, and Yi Zeng.
\newblock Backpropagation with biologically plausible spatiotemporal adjustment
  for training deep spiking neural networks.
\newblock \emph{Patterns}, 3\penalty0 (6), 2022.

\bibitem[Shen et~al.(2023{\natexlab{a}})Shen, Zhao, Dong, and
  Zeng]{shen2023brain}
Guobin Shen, Dongcheng Zhao, Yiting Dong, and Yi Zeng.
\newblock Brain-inspired neural circuit evolution for spiking neural networks.
\newblock \emph{Proceedings of the National Academy of Sciences}, 120\penalty0
  (39):\penalty0 e2218173120, 2023{\natexlab{a}}.

\bibitem[Shen et~al.(2023{\natexlab{b}})Shen, Zhao, and
  Zeng]{shen2023exploiting}
Guobin Shen, Dongcheng Zhao, and Yi Zeng.
\newblock Exploiting high performance spiking neural networks with efficient
  spiking patterns.
\newblock \emph{arXiv preprint arXiv:2301.12356}, 2023{\natexlab{b}}.

\bibitem[Wu et~al.(2021)Wu, Xu, Han, Zhou, Zhang, Li, and
  Tan]{wu2021progressive}
Jibin Wu, Chenglin Xu, Xiao Han, Daquan Zhou, Malu Zhang, Haizhou Li, and
  Kay~Chen Tan.
\newblock Progressive tandem learning for pattern recognition with deep spiking
  neural networks.
\newblock \emph{IEEE Transactions on Pattern Analysis and Machine
  Intelligence}, 44\penalty0 (11):\penalty0 7824--7840, 2021.

\bibitem[Yao et~al.(2023)Yao, Hu, Zhou, Yuan, Tian, Xu, and Li]{yao2023spike}
Man Yao, Jiakui Hu, Zhaokun Zhou, Li Yuan, Yonghong Tian, Bo Xu, and Guoqi Li.
\newblock Spike-driven transformer.
\newblock \emph{arXiv preprint arXiv:2307.01694}, 2023.

\bibitem[Zeng et~al.(2023)Zeng, Zhao, Zhao, Shen, Dong, Lu, Zhang, Sun, Liang,
  Zhao, et~al.]{zeng2023braincog}
Yi Zeng, Dongcheng Zhao, Feifei Zhao, Guobin Shen, Yiting Dong, Enmeng Lu, Qian
  Zhang, Yinqian Sun, Qian Liang, Yuxuan Zhao, et~al.
\newblock Braincog: A spiking neural network based, brain-inspired cognitive
  intelligence engine for brain-inspired ai and brain simulation.
\newblock \emph{Patterns}, 4\penalty0 (8), 2023.

\bibitem[Zhang et~al.(2018)Zhang, Qu, Belatreche, Chen, and
  Yi]{zhang2018highly}
Malu Zhang, Hong Qu, Ammar Belatreche, Yi Chen, and Zhang Yi.
\newblock A highly effective and robust membrane potential-driven supervised
  learning method for spiking neurons.
\newblock \emph{IEEE transactions on neural networks and learning systems},
  30\penalty0 (1):\penalty0 123--137, 2018.

\bibitem[Zhang et~al.(2021)Zhang, Wang, Wu, Belatreche, Amornpaisannon, Zhang,
  Miriyala, Qu, Chua, Carlson, et~al.]{zhang2021rectified}
Malu Zhang, Jiadong Wang, Jibin Wu, Ammar Belatreche, Burin Amornpaisannon,
  Zhixuan Zhang, Venkata Pavan~Kumar Miriyala, Hong Qu, Yansong Chua, Trevor~E
  Carlson, et~al.
\newblock Rectified linear postsynaptic potential function for backpropagation
  in deep spiking neural networks.
\newblock \emph{IEEE transactions on neural networks and learning systems},
  33\penalty0 (5):\penalty0 1947--1958, 2021.

\bibitem[Zhang et~al.(2022)Zhang, Zhang, and Lew]{zhang2022pokebnn}
Yichi Zhang, Zhiru Zhang, and Lukasz Lew.
\newblock Pokebnn: A binary pursuit of lightweight accuracy.
\newblock In \emph{Proceedings of the IEEE/CVF Conference on Computer Vision
  and Pattern Recognition}, pages 12475--12485, 2022.

\bibitem[Zhao et~al.(2023)Zhao, Shen, Dong, Li, and Zeng]{zhao2023improving}
Dongcheng Zhao, Guobin Shen, Yiting Dong, Yang Li, and Yi Zeng.
\newblock Improving stability and performance of spiking neural networks
  through enhancing temporal consistency.
\newblock \emph{arXiv preprint arXiv:2305.14174}, 2023.

\bibitem[Zheng et~al.(2021)Zheng, Wu, Deng, Hu, and Li]{zheng2021going}
Hanle Zheng, Yujie Wu, Lei Deng, Yifan Hu, and Guoqi Li.
\newblock Going deeper with directly-trained larger spiking neural networks.
\newblock In \emph{Proceedings of the AAAI conference on artificial
  intelligence}, pages 11062--11070, 2021.

\bibitem[Zhou et~al.(2016)Zhou, Wu, Ni, Zhou, Wen, and Zou]{zhou2016dorefa}
Shuchang Zhou, Yuxin Wu, Zekun Ni, Xinyu Zhou, He Wen, and Yuheng Zou.
\newblock Dorefa-net: Training low bitwidth convolutional neural networks with
  low bitwidth gradients.
\newblock \emph{arXiv preprint arXiv:1606.06160}, 2016.

\bibitem[Zhou et~al.(2022)Zhou, Zhu, He, Wang, Yan, Tian, and
  Yuan]{zhou2022spikformer}
Zhaokun Zhou, Yuesheng Zhu, Chao He, Yaowei Wang, Shuicheng Yan, Yonghong Tian,
  and Li Yuan.
\newblock Spikformer: When spiking neural network meets transformer.
\newblock \emph{arXiv preprint arXiv:2209.15425}, 2022.

\bibitem[Zhu et~al.(2023)Zhu, Zhao, and Eshraghian]{zhu2023spikegpt}
Rui-Jie Zhu, Qihang Zhao, and Jason~K Eshraghian.
\newblock Spikegpt: Generative pre-trained language model with spiking neural
  networks.
\newblock \emph{arXiv preprint arXiv:2302.13939}, 2023.

\end{thebibliography}
}

\clearpage
\setcounter{page}{1}
\maketitlesupplementary

\section{Experimental Details}
\label{suppl:data}
\subsection{Static Datasets} 

In order to verify the efficiency of QSNN, we performed a series of comprehensive evaluations on several datasets. We describe these datasets in more detail below.


\textbf{ImageNet}~\cite{deng2009imagenet} With over a million categorized images spread across $1,000$ diverse categories, ImageNet stands as a colossal and pivotal dataset. The images in ImageNet, which cover a wide spectrum from tangible items to abstract notions, demonstrate considerable variations in attributes like scale, pose, and illumination. Consequently, it provides a rigorous testing ground for diverse computer vision tasks and serves as an industry-standard benchmark.

\textbf{CIFAR}~\cite{krizhevsky2009learning} The CIFAR datasets, particularly CIFAR10 and CIFAR100, are quintessential datasets in the field of machine learning, commonly employed for assessing image classification algorithms. CIFAR10 comprises $60,000$ $32 \times 32$ color images, evenly distributed across 10 distinct categories, with each category containing $6,000$ images. In contrast, CIFAR100 maintains the same overall number of images and resolution but is divided into 100 categories, each containing 600 images, offering a finer granularity of classification. 

\subsection{Neuromorphic Datasets}

Neuromorphic datasets provide a benchmark for assessing algorithms tailored to neuromorphic hardware, which often deal with spiking neural networks (SNNs) and event-driven data. Such datasets capture real-world, dynamic visual information in a format suitable for SNNs, emphasizing the importance of time and asynchronous events in information processing.

\textbf{CIFAR10-DVS}~\cite{li2017cifar10} The CIFAR10-DVS dataset is a neuromorphic version of the popular CIFAR10 dataset. Instead of static images, CIFAR10-DVS provides sequences of events generated by a Dynamic Vision Sensor (DVS), a type of neuromorphic camera that only captures pixel-level brightness changes, making it more power-efficient and better suited for real-time applications.

\textbf{DVS-Gesture}~\cite{amir2017low} The DVS-Gesture dataset is a benchmark collection in the field of neuromorphic vision, designed for the task of gesture recognition. The DVS-Gesture dataset includes a variety of hand gestures from multiple subjects, performed under different lighting conditions, and from various angles, challenging the robustness of spiking neural network (SNN) models in recognizing and classifying dynamic patterns.

\textbf{N-Caltech101}~\cite{orchard2015converting} The N-Caltech101 dataset is a neuromorphic version of the well-known Caltech101 dataset, which has been converted using a DVS camera to create event-based vision data. Unlike the original dataset with static images, N-Caltech101 provides a sequence of events as each image is presented to the sensor, capturing the temporal aspect of visual perception. This dataset contains the same categories as the original Caltech101, encompassing a wide range of objects such as animals, vehicles, and everyday items, thus providing a comprehensive challenge for testing the effectiveness of SNNs in processing and classifying neuromorphic vision data.

\subsection{Experimental Settings}


Following the framework defined by Zhou et al.~\cite{zhou2022spikformer} for consistency and comparability, we set the input dimensions for the ImageNet dataset to $224 \times 224$. For other static image datasets, their native image dimensions were retained. The neural morphology datasets underwent a resizing process, adjusting the event streams to $48 \times 48$ to streamline computational demands. We used a batch size of 128 and selected the AdamW optimizer~\cite{loshchilov2017decoupled}. Training involved 310 epochs for the ImageNet dataset, whereas for other datasets, 400 epochs were deemed sufficient. We initialized our models with a learning rate of $0.0005$.

Our experiments spanned various neural network architectures, prominently SEW-ResNet~\cite{fang2021deep} and SpikFormer~\cite{zhou2022spikformer}. Although the architectural blueprints of our models mirrored the original specifications, we took the liberty of introducing modifications to accentuate their event-driven nature. This involved imposing bit width restrictions on the output of SEW-ResNet's blocks and SpikFormer's attention matrices. Our thorough evaluations then centered on the performances of S-ACE and NS-ACE for these models across our selected datasets.

\subsection{Additional Evaluation}
\label{suppl:cifar}

In Tab.~\ref{table:cifar_result_suppl}, various methods and configurations showcase their performance on both CIFAR10 and CIFAR100. Burst\cite{shen2023exploiting}, although demonstrating impressive performance across different configurations, retains a relatively large number of parameters due to its limited consideration of pulse patterns and its lack of quantization for model weights. Building upon Spikiformer~\cite{zhou2022spikformer}, our further experiments reveal that models with multiple pulse patterns can better utilize the bit budget. When combined with weight quantization, we managed to achieve an accuracy of $96.84\%$ on CIFAR10 and $80.13\%$ on CIFAR100 with just 1/4 of the original model parameters and 1/5 of the S-ACE. These results indicate a performance improvement of $1.33\%$ and $1.92\%$ respectively over the non-optimized methods.

From the neuromorphic results in Tab.~\ref{table:dvs_result}, it's evident that as we modify the allocation between the spike patterns and simulation steps (S and T), there are significant performance trade-offs. 
The most notable decline in performance is observed for the DVSC10 dataset when adjusting the S/T ratio from 1/16 to 8/2, indicating a substantial sensitivity to the bit allocation strategy. This is further accentuated in the 1/16/1 configuration, leading to a staggering decrease of $44\%$ in accuracy. 
However, the impact of weight bit-width on model performance is not particularly pronounced. This might be attributed to the relatively small dataset size of the neural morphology dataset, where weight bit-width may not be the predominant factor influencing model performance.

Building upon our previous discussions and the observed performance on the CIFAR dataset, these results further underscore the intricacy of employing spiking neural networks on the neural morphology dataset. The judicious selection and allocation of bits are of paramount importance when dealing with neural morphology data characterized by temporal dynamics.

\begin{table*}[t]
    \caption{Additional evaluation on CIFAR10/100. $^\dagger$ represent results reproduced with the same experimental setup.}
    \begin{center}
        \tabcolsep=0.16cm
        \begin{tabular}{ccccccc}
            \toprule
            Methods   & Architecture   & \tabincell{c}{Assignment \\ W / S / T} &  \tabincell{c}{Params \\ (M)} & \tabincell{c}{S-ACE \\ (G)}  & \tabincell{c}{Accuracy \\ CIFAR10}  & \tabincell{c}{Accuracy \\ CIFAR100}\\
            \midrule
            PLIF~\cite{fang2021incorporating}      & CIFARNet-Fang     & 16 / 1 / 8   & 0.58  & 3.57 & 93.50 & 74.36$^\dagger$ \\
            Diet-SNN~\cite{rathi2020enabling}            & ResNet-20         & 16 / 1 / 5   & 0.27  & 3.32 & 92.54 & 64.07$^\dagger$ \\
            tdBN~\cite{zheng2021going}         & ResNet-19         & 16 / 1 / 4   & 12.63 & 139  & 92.92 & 70.86 \\
            TET~\cite{deng2022temporal}            & ResNet-19         & 16 / 1 / 4   & 12.63 & 139  & 94.44 & 74.47 \\
                                                                    \cline{2-7}
            \multirow{4}{*}{Burst~\cite{shen2023exploiting}}          
                                                                    & ResNet-19         & 16 / 2 / 1   & 12.63 & 69.6 & 95.94 & 77.86 \\
                                                                    & ResNet-19         & 16 / 2 / 2   & 12.63 & 139  & 96.01 & 78.04 \\
                                                                    & ResNet-19         & 16 / 2 / 4   & 12.63 & 278  & 96.21 & 78.12 \\
                                                                    & ResNet-19         & 16 / 2 / 6   & 12.63 & 417  & 96.32 & 78.31 \\
            \hline
            \multirow{3}{*}{Spikformer~\cite{zhou2022spikformer}}
                                                                    & Spikformer-4-256  & 16 / 1 / 4   & 4.15  & 26.28 & 93.94 & 75.96  \\ 
                                                                    & Spikformer-2-384  & 16 / 1 / 4   & 4.15  & 44.50 & 94.80 & 76.95  \\ 
                                                                    & Spikformer-4-384  & 16 / 1 / 4   & 4.15  & 59.10 & 95.51 & 78.21  \\ 
            \hline 
            \multirow{12}{*}{\tabincell{c}{Quantized \\ Spikformer}}           
                                                                    & Spikformer-4-384  & 1 / 1 / 1    & 0.26  & 0.77 & 90.48 & 70.11  \\          
                                                                    \cline{2-7}      
                                                                    & Spikformer-4-384  & 1 / 4 / 1    & 0.26  & 3.08 & 95.00 & 76.90  \\     
                                                                    & Spikformer-4-384  & 1 / 2 / 2    & 0.52  & 3.08 & 94.43 & 75.91  \\   
                                                                    & Spikformer-4-384  & 1 / 1 / 4    & 0.26  & 3.69 & 93.91 & 74.13  \\    
                                                                    \cline{2-7}
                                                                    & Spikformer-4-384  & 2 / 2 / 1    & 0.52  & 3.08 & 95.41 & 76.67  \\     
                                                                    & Spikformer-4-384  & 2 / 1 / 2    & 0.52  & 3.09 & 93.56 & 75.91  \\    
                                                                    \cline{2-7}             
                                                                    & Spikformer-4-384  & 4 / 1 / 1    & 1.04  & 3.09 & 94.51 & 74.61  \\            
                                                                    & Spikformer-4-384  & 4 / 4 / 1    & 1.04  & 12.32 & 96.84 & 80.13  \\      
                                                                    & Spikformer-4-384  & 4 / 2 / 2    & 1.04  & 12.33 & 96.50 & 80.71  \\      
                                                                    & Spikformer-4-384  & 4 / 1 / 4    & 1.04  & 14.77 & 95.94 & 78.77  \\      
                                                                    \cline{2-7}
                                                                    & Spikformer-4-384  & 8 / 1 / 2    & 2.08  & 12.35 & 95.55 & 77.72  \\      
                                                                    & Spikformer-4-384  & 8 / 2 / 1    & 2.08  & 12.33 & 96.29 & 80.00  \\        
            \hline   
        \end{tabular}
    \end{center}
    \label{table:cifar_result_suppl}
    \vspace{-4mm}
\end{table*}


\section{Hardware Implementation}
\label{suppl:hardware}


\end{document}